\pdfoutput=1
\documentclass[11pt]{article}
\usepackage{booktabs}
 \usepackage{EMNLP2022}
\usepackage{comment}

\usepackage{times}
\usepackage{latexsym}

\usepackage{tabularx}
\usepackage{makecell}
\usepackage{multirow}
\usepackage{xspace}
\usepackage{scrextend}
\usepackage[inline]{enumitem}
\usepackage{tikz}
\usepackage{float}
\usepackage[draft]{todo}

\usepackage[T1]{fontenc}
\usepackage[utf8]{inputenc}

\usepackage{microtype}

\usepackage{inconsolata}

\newcounter{notecounter}
\newcommand{\enotesoff}{\long\gdef\enote##1##2{}}
\newcommand{\enoteson}{\long\gdef\enote##1##2{{
\stepcounter{notecounter}
{\large\bf
\hspace{0cm}\arabic{notecounter} $<<<$ ##1: ##2
$>>>$\hspace{1cm}}}}}

\enoteson
 \enotesoff

\def\ourmethod{GLP\xspace}
\def\ourmethodone{GLP-B\xspace}
\def\ourmethodtwo{GLP-SL\xspace}

\title{Graph-Based Multilingual Label Propagation for Low-Resource Part-of-Speech Tagging}

\author{Ayyoob Imani \textsuperscript{*1}, Silvia Severini\textsuperscript{*1}, \\ \textbf{Masoud Jalili Sabet}\textsuperscript{1},  \textbf{François Yvon}\textsuperscript{2}, \textbf{Hinrich Schütze}\textsuperscript{1} \\
	\textsuperscript{1}Center for Information and Language Processing (CIS), LMU Munich, Germany \\
	\textsuperscript{2}Université Paris-Saclay, CNRS, LISN, France \\
	\texttt{\{ayyoob, silvia, masoud\}@cis.lmu.de,} 
	\texttt{francois.yvon@limsi.fr} \\}

\begin{document}
\maketitle
\def\thefootnote{*}\footnotetext{Equal contribution.}\def\thefootnote{\arabic{footnote}}
\begin{abstract}
Part-of-Speech (POS) tagging is an important component of the NLP pipeline, 
but many
low-resource languages lack labeled data for training.  An
established method for training a POS tagger in such a
scenario is to create a labeled training set by transferring
from high-resource languages. In this paper, we propose a
novel method for transferring labels from multiple high-resource
source to low-resource target languages.
We formalize POS tag projection as graph-based label
propagation.  Given translations of a sentence in multiple
languages, we create a graph with words as nodes and
alignment links as edges by aligning words for all language
pairs. We then propagate node labels from source to target
using a Graph Neural Network 
augmented with
transformer layers. 
We show that our propagation creates training sets
that allow us to train
POS taggers for a diverse set of languages.
When combined with enhanced contextualized embeddings,
our method achieves a new 
state-of-the-art
for unsupervised POS tagging of low 
resource languages.

\end{abstract}

\section{Introduction} %(take 3 = mix 1 and 2)
\begin{figure}[!ht] \centering 
		\includegraphics[width=1.0\linewidth, trim={0.9cm 6.5cm 8.8cm 1.9cm }, clip]{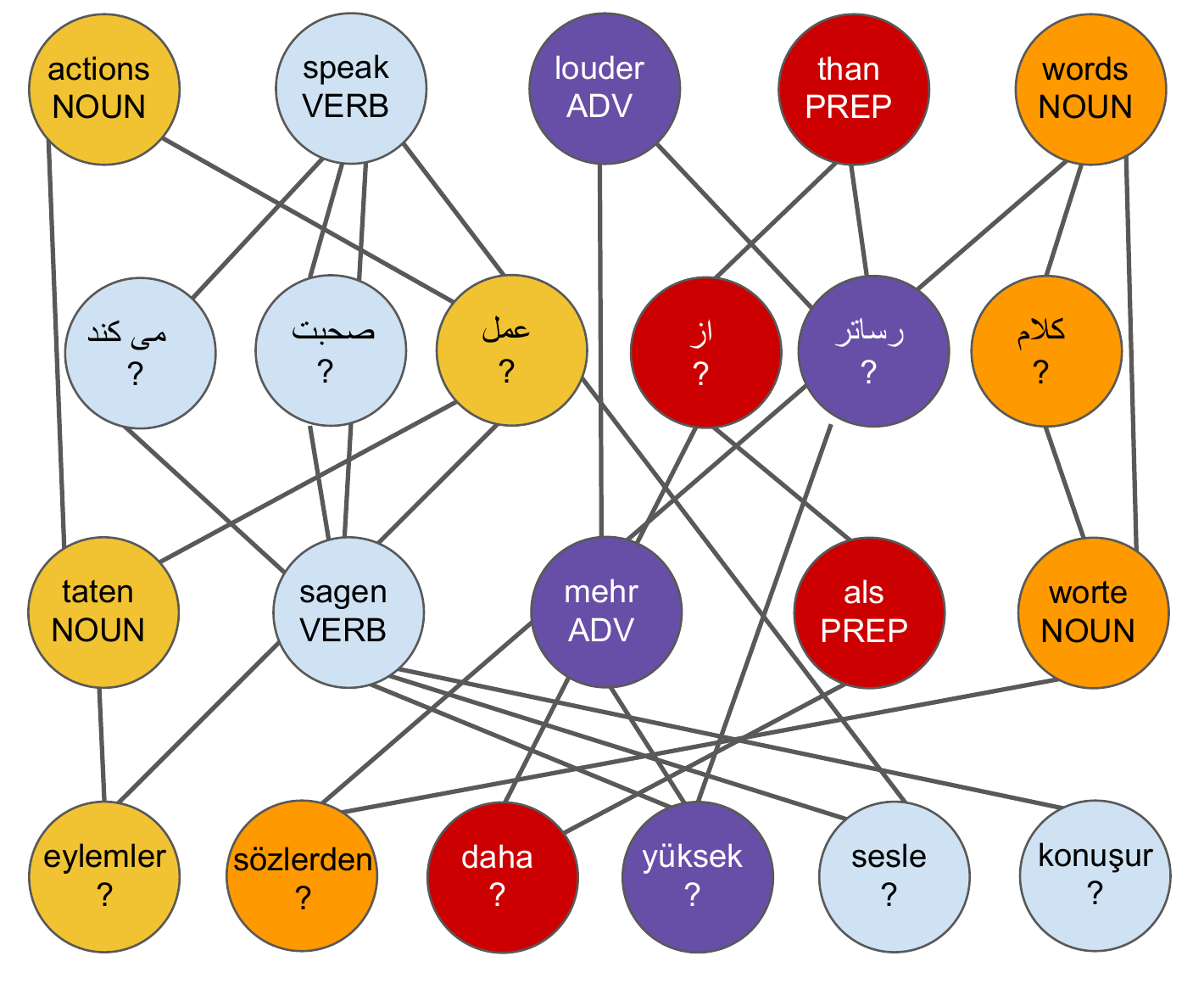} 
	\caption{The sentence ``actions speak louder than words'' in English and its translations to 
	Persian, German, and Turkish, aligned at the word level.
	The POS tags for high-resource English and German
	are known. We use a GNN to exploit
        this graph structure and compute POS tags for low-resource Persian and Turkish.
		}
	\label{fig:example}
\end{figure}
In many applications,
Part-of-Speech (POS) tagging is an important
part of the NLP pipeline.
In recent years, high-accuracy POS taggers have been
developed owing to advances in machine learning methods that
combine pretraining on large unlabeled corpora  and
supervised fine-tuning
on well-curated annotated
datasets. This methodology only applies to a handful of
high-resource (HR) languages for which the necessary
training data exists, leaving behind the majority of low-resource (LR) languages.
When training resources are scarce, an established method for training POS taggers is to automatically generate 
the training data via cross-lingual transfer
\citep{yarowsky-ngai-2001-inducing,fossum-abney-2005-automatically,agic2016multilingual,eskander-etal-2020-unsupervised}. 
Typically, POS annotations are 
projected
through alignment links from the HR source  to the LR target  of a word aligned parallel corpus.  

In this paper, we propose \textbf{\ourmethod} (\textbf{G}raph \textbf{L}abel \textbf{P}ropagation), a novel method for transferring labels simultaneously 
\emph{from multiple high-resource source languages to multiple low-resource target languages}.
We formalize POS tag projection as graph-based label propagation. Given translations of a sentence in multiple languages, we create a graph with words as nodes and
alignment links as edges by aligning words for all language pairs. 
We then propagate POS labels associated with source language nodes 
to target language nodes, using a 
label propagation model that is formalized as a
Graph Neural Network (GNN) \citep{scarselli2008graph}.
Nodes are represented by a diverse set of features that describe both linguistic properties 
and graph structural information. In a second step, we additionally employ self-learning
to obtain reliable training instances in the target languages.

Our approach is based on \emph{multiparallel corpora}, meaning that the translation
of each sentence is available in more than two languages.
We exploit the Parallel Bible Corpus (PBC) of 
\citet{mayer-cysouw-2014-creating},\footnote{We do not use
PBC-specific features. Thus,
our work is in principle applicable to any multiparallel corpus.}
a multiparallel corpus covering more than 1000 languages, 
many of which are extremely low-resource, 
by which we mean
that only a tiny amount of unlabeled data is available or
that no language technologies exist for them at all \citep{joshi-etal-2020-state}.

We evaluate our method on a diverse set of low-resource languages from
multiple language families, including four 
languages not covered by pretrained language models (PLMs).
We train POS tagging models
for these languages and evaluate them against references from the Universal
Dependencies corpus \citep{UD}. 
We compare the results of our method against multiple state-of-the-art (SOTA)
cross-lingual unsupervised and semisupervised POS taggers employing different
approaches like annotation projection and zero-shot transfer.
Our experiments
highlight the benefits of our new transfer and self-learning
methods; crucially, they show that reasonably accurate POS
taggers can be bootstrapped without any 
annotated data
for a
diverse set of
low-resource languages, establishing a new SOTA
for high-resource-to-low-resource cross-lingual POS
transfer. 
We also assess the quality of the projected
annotations with respect to ``silver'' references and perform an
ablation study.
To summarize, our contributions 
are:\footnote{Our code, data, and trained models are available at \url{https://github.com/ayyoobimani/GLP-POS}}
\begin{itemize}
    \item We formalize annotation projection as
    graph-based label propagation and introduce two new POS
    annotation projection models, \ourmethodone{} (\ourmethod-Base)
    and \ourmethodtwo (\ourmethod-SelfLearning).
    \item We evaluate \ourmethodone{}
    and \ourmethodtwo{}
    on 17 low-resource languages, including 4 languages not covered
    by large PLMs.
	\item By comparing our method with various supervised, semisupervised, 
			and PLM-based approaches
			for POS tagging of low-resource languages, we establish a new SOTA
			for unsupervised POS tagging.
\end{itemize}    

\section{Related work \label{sec:related}}

\paragraph{POS tagging}
Part of Speech 
tagging aims to assign each word the proper syntactic tag in context \citep{manning99foundations}. 
For high-resource languages, for which large labeled training
sets are available,
high-accuracy POS tagging is achieved 
through supervised learning \cite{kondratyuk-straka-2019-75, tsai2019small}. 

\paragraph{Zero-shot transfer}
In low-resource settings, 
one
approach is to use
cross-lingual transfer thanks to pretrained multilingual
representations, thereby enabling zero-shot POS tagging.
\citet{kondratyuk-straka-2019-75} analyze the few-shot and
zero-shot performance of
mBERT \citep{devlin-etal-2019-bert} fine-tuning on POS tagging.
We include this approach in our set of baselines below.
\citet{ebrahimi-kann-2021-adapt}
and \citet{wang-etal-2022-expanding} analyze zero-shot POS tagging
performance of
XLM-RoBERTa \citep{conneau-etal-2020-unsupervised}
and propose complementary methods such as
continued pretraining, vocabulary expansion and adapter modules for
better performance.
We show that combining \ourmethod{} with \citet{wang-etal-2022-expanding}'s embeddings further
improves our base performance.

\paragraph{Annotation projection} Annotation projection is another approach to annotating 
low-resource languages.
\citet{yarowsky-ngai-2001-inducing} first proposed 
projecting annotation labels across languages, exploiting parallel corpora and word alignment.
To reduce systematic transfer errors, {\citet{fossum-abney-2005-automatically} extended this by projecting from multiple source languages.
\citet{agic-etal-2015-bit} and  \citet{agic2016multilingual}
exploit multilingual transfer setups to bootstrap POS
taggers for low-resource languages starting from a parallel corpus and taggers and parsers for high-resource languages.
Other works project labels by leveraging token and type-level constraints \citep{tackstrom-etal-2013-token,buys-botha-2016-cross,eskander-etal-2020-unsupervised}. The latter study notably proposes an
unsupervised method for selecting training instances via cross-lingual projection and trains POS taggers exploiting contextualized word embeddings, affix embeddings and hierarchical Brown clusters \citep{brown-etal-1992-class}. This approach is also used as a baseline below.

Semi-supervised approaches have been proposed to mitigate the noise of projecting between languages. This can be achieved with auxiliary lexical resources \citep{tackstrom-etal-2013-token,ganchev-das-2013-cross,wisniewski2014cross,
li-etal-2012-wiki} that guide unsupervised learning or act as
an additional training signal
\citep{plank-agic-2018-distant}.
Other works combine manual and projected
annotations \citep{garrette-baldridge-2013-learning,fang-cohn-2016-learning}.
We outperform prior works without the use of additional resources
such as dictionaries or annotations.

\paragraph{Graph Neural Networks}
Many natural and real-life structures 
like physical systems,
social networks \& interactions, and molecular fingerprints
have a graph structure \cite{liu2020introduction}. 
Graph neural networks have been successfully used to model them.
Applications include social spammer detection \cite{Wu_Lian_Xu_Wu_Chen_2020},
learning molecular fingerprints \cite{duvenaud2015convolutional} and 
human motion prediction \cite{li2020dynamic}. Recently,
GNNs have been adopted for NLP tasks such as text 
classification \cite{peng2018large}, sequence labeling 
\cite{zhang-etal-2018-sentence, marcheggiani-titov-2017-encoding},
neural machine translation \cite{bastings-etal-2017-graph,beck-etal-2018-graph},
and alignment link prediction \cite{imani-etal-2022-graph}. As far as
we know, our work is the first to
formalize the annotation projection problem as graph-based label propagation.

\paragraph{Multiparallel corpora}
A multiparallel corpus provides the translations of a source text
in more than two languages. A few such corpora 
\cite{agic-vulic-2019-jw300, mayer-cysouw-2014-creating, TIEDEMANN12.463}
 provide sentence-level aligned 
text for hundreds or thousands of languages; for many of
these languages
only a tiny amount of digitized content is available \cite{joshi-etal-2020-state}.
Although the amount of text found in existing multiparallel corpora 
is far less than in monolingual corpora, we believe that they can serve
as cross-lingual bridges, with which
effective representation for low-resource languages can be derived. Highly multiparallel corpora have been used 
for expanding pretrained models to more languages \cite{ebrahimi-kann-2021-adapt,wang-etal-2022-expanding}, 
word alignment improvement and visualization \cite{imanigooghari-etal-2021-parcoure,imani-etal-2022-graph},
embedding learning \cite{dufter2018embedding},  and annotation projection \cite{agic2015if, severini2022towards}.

\section{Method \label{sec:method}}
We now introduce our \textit{Graph Label Propagation}
(\ourmethod{}) method, which
formalizes the problem of annotation projection as graph-based label propagation.
We first describe the graph structure, then the features associated with each node,
and finally the architecture of our model.

\subsection{Problem formalization}
\begin{figure}[!t] \centering 
	\includegraphics[width=1\linewidth, trim = {2cm 9.5cm 4.9cm 5.2cm}, clip]{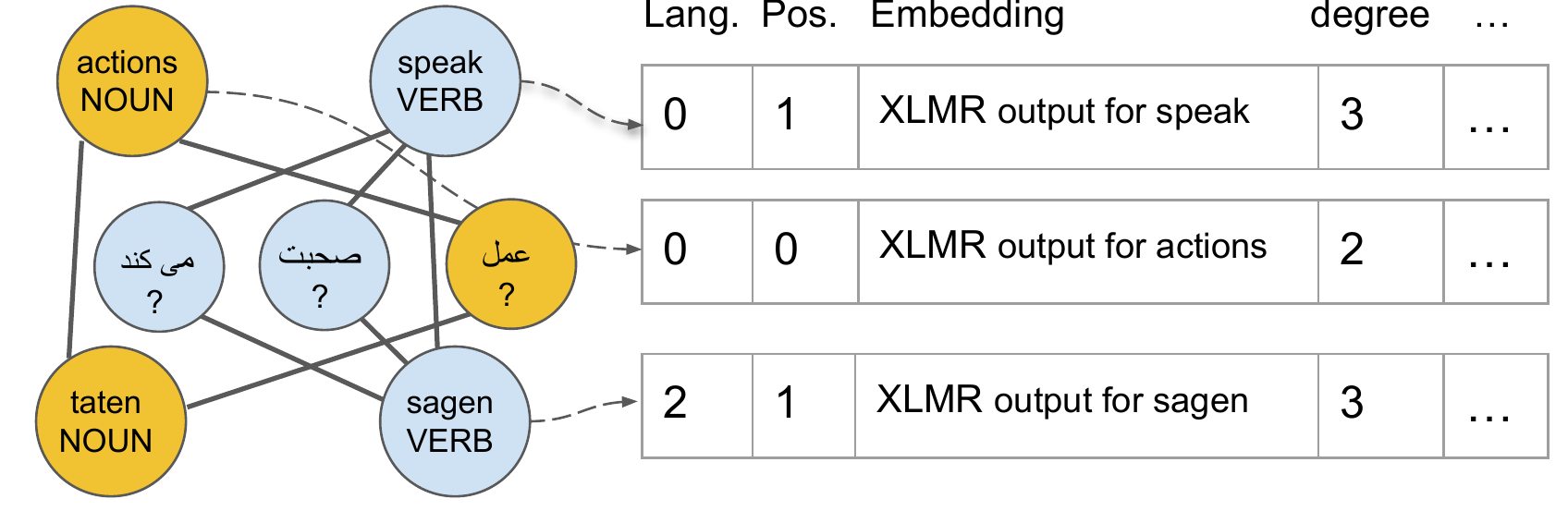} 
	\caption{An example of how we represent nodes of an alignment graph using features
	for a part of the graph in Figure \ref{fig:example}.}
	\label{fig:features}
\end{figure}

The  \textit{multilingual alignment graph} (MAG) of a sentence is formalized as follows.
Each sentence $\sigma$ in our multiparallel corpus exists in a set $L$ of 
languages.\footnote{$|L|$ might be different for different sentences.}
 $L$ contains both
high-resource source languages (in $L_s$) and low-resource target languages (in $L_t$) with
$L_s\cup L_t=L$. Each word in these $|L|$ versions of $\sigma$ will constitute a node in our graph.
We first automatically annotate the text in all the source languages using pre-existing taggers:
these POS tags are node labels; they are only known for languages in $L_s$, unknown otherwise.
We then use Eflomal \citep{Ostling2016efmaral}, an unsupervised word alignment tool
to compute alignment links for all $\frac{|L|*(|L|-1)}{2}$ language pairs:
these links define the edges of our MAG.
Figure \ref{fig:example} displays an example MAG for four languages, with
English and German as sources and Turkish and Persian as targets. 
Note that both the word alignments and the node labels are noisy, since we
do not use gold data but statistical methods to generate
them.

\subsection{Features \label{ssec:features}}
To train graph neural networks, we represent each node using a set of features \cite{duong2019node}.
In Figure~\ref{fig:features}, you see a simple illustration of how nodes are 
represented using a feature vector. The graph in this figure is part of the original 
graph in Figure~\ref{fig:example}.
Two types of features are considered:
features that represent the inherent meaning of a node/word (word representation features)
and features that describe the position of a node within the graph (graph structural features). 
Node representation features consist of:
XLM-R \cite{conneau-etal-2020-unsupervised} embeddings, the node's language
and its position within the sentence. 
Since XLM-R embeddings are not available for all
languages,
we alternatively experiment with static word embeddings created
using \citet{levy-etal-2017-strong}'s sentence-ID method, which we train on PBC.
Our graph structural features are similar
to \citet{imani-etal-2022-graph}'s work on
link prediction.
They include five centrality features:
\textit{degree, closeness} \citep{freeman1978centrality},
\textit{betweenness} \citep{brandes2001faster}, \textit{load} \citep{newman2001scientific},
and \textit{harmonic centrality} \citep{boldi2014axioms}. Each of these features describes
the node's position within the graph from a different perspective. For example, 
\textit{degree} is the number of neighbors of the node and 
\textit{harmonic centrality} measures how important/influential a node is.
They also include two community features corresponding to
the ID of the node's communities computed respectively with the 
greedy modularity community detection method of \citet{clauset2004finding} and 
the label propagation algorithm of \citet{cordasco2010community}.
These two methods detect communities of nodes such that there are
many links within the communities and only a few between
them. 

\begin{figure*}
	\centering 
		\includegraphics[width=0.99\textwidth, trim={0.9cm 10.0cm 2.6cm 4.0cm}]{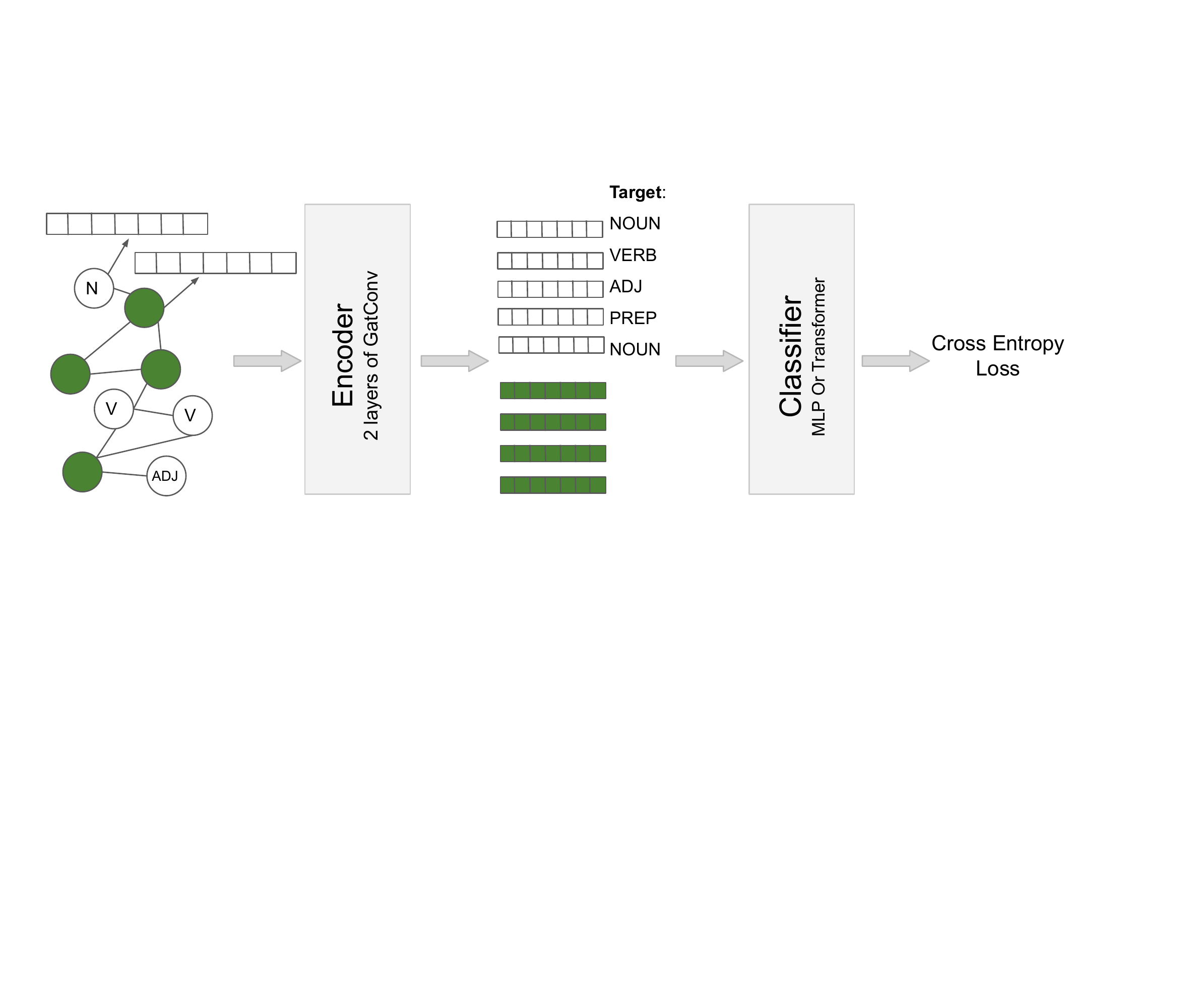} 
	\caption{The architecture of \ourmethod{} (Graph
	Label Propagation). Source nodes are in white, 
    target nodes  in green. For training, 
    we first feed the  alignment
    graph of a sentence to the encoder to compute a representation for each node. 
    Next we feed the representations of the source nodes
to the classifier. The training objective is cross entropy
	on prediction of POS tags. Note that we know the 
    POS tags of the source nodes.
	After training, the model can generalize the POS tag
	prediction to target nodes. 
	}
	\label{fig:gnn_architecture}
\end{figure*}

\subsection{\ourmethod{} architecture}
Figure~\ref{fig:gnn_architecture} displays the architecture of our \ourmethod{} model; white nodes are  for the source (= training) languages
and green nodes for the target languages.
The model has two parts: the GNN-based \emph{encoder} turns
the alignment graph into node representations and the \emph{classifier}
learns to label nodes based on these representations.
The network is trained to reproduce POS tags for each source node;
it is then used to predict the unknown tags for target nodes.

The encoder has two GATConv layers \cite{velickovic2018graph}:
given a graph with $M$ nodes represented as $\mathbf{x}_1, \mathbf{x}_2, ..., \mathbf{x}_M$, 
with respective neighborhoods $\mathcal{N}(1), \mathcal{N}(2), ..., \mathcal{N}(M)$, 
a GATConv layer computes a new representation $\mathbf{x}^{\prime}_i$ for each node as:
\begin{equation}
	\mathbf{x}^{\prime}_i = 
			\sum_{j \in \mathcal{N}(i) \cup \{ i \}} \alpha_{i,j}\mathbf{W}\mathbf{x}_{j},
\end{equation}
where $\mathbf{W}$ is a learnable weight matrix. $\alpha_{i,j}$
measures how much node $i$ ``attends'' to node $j$ as follows:
\begin{eqnarray}
\nonumber	\alpha_{i,j} =
	\frac{
	\exp\left(g\left(\mathbf{a}^{\top}
	[\mathbf{W}\mathbf{x}_i \, \Vert \, \mathbf{W}\mathbf{x}_j]
	\right)\right)}
	{\sum_{k \in \mathcal{N}(i) \cup \{ i \}}
	\exp\left(g\left(\mathbf{a}^{\top}
	[\mathbf{W}\mathbf{x}_i \, \Vert \, \mathbf{W}\mathbf{x}_k]
	\right)\right)} 
\end{eqnarray}
where $\Vert$ stands for concatenation, $g$ is the LeakyReLU \cite{maas2013rectifier}, and 
$\mathbf{a}$ is a weight vector. As neighborhoods only use alignment links, the representation
of a node is only influenced by nodes in other languages.
Also note that both source and target nodes 
are fed to the encoder.

We train two \ourmethod{} models: 
\ourmethod{}-Base (\ourmethodone) and 
\ourmethod{}-SelfLearning
(\ourmethodtwo).
The first one is the basic GNN architecture.
It tags a token based on the other languages only,
i.e.\, it makes no use of the sequence information of the
current token in its own language.
The second additionally employs self-learning and is given access to
the local context of each token in its own language.

\paragraph{\ourmethodone{}}
uses a multi-layer perceptron as
classifier. We feed the node
representations to the classifier and train the 
model end-to-end. We can only do this for source nodes
since we have no training data for the
target languages. 

\paragraph{\ourmethodtwo{}}
additionally employs self-learning and a better
classifier. Self-learning takes advantage of node labels predicted by
\ourmethodone{} in the first step: when the prediction
confidence exceeds a threshold $\gamma$, these labels are deemed
correct and the corresponding nodes are considered when
training the classifier. \ourmethodtwo{} uses a Transformer architecture
to predict POS tags. The Transformer input consists of all translations
of a sentence, where words are represented as GNN node embeddings. Each
embedding is the concatenation of input ($x_i$) and output
representations ($x^{\prime}_i$) of the corresponding node in
the GNN. In addition to the information available
from neighbor nodes in \emph{other} languages,
the Transformer can attend to other words of the sentence in
the \emph{same} language, some of which may already be (automatically)
labeled. This is very different from the training of
\ourmethodone{}, where the POS of words of the same language were either
all known (for source languages) or all unknown (for target languages),
and explains why we resorted to a simpler classifier in the first stage.

Similarly to 
\citet{eskander-etal-2020-unsupervised} and \citet{agic2016multilingual},
\ourmethodtwo{} uses type-level information:
for each word type, we create a tag distribution by accumulating counts of 
the number of times each tag was assigned.
For source words, we use the training data to estimate the distribution.
For target words,  we use the predictions of  \ourmethodone{}
on PBC.

\subsection{Neural POS tagger}

We use the noisy labeled data, generated by \ourmethodone{} or \ourmethodtwo{},
to train monolingual neural POS taggers.
Each model is a
Bi-LSTM
(Bidirectional Long Short-Term Memory, \citep{hochreiter1997long}  with XLM-RoBERTa embeddings \citep{conneau-etal-2020-unsupervised}.
The input is a sentence labeled by \ourmethodone{} or \ourmethodtwo{}.
A token  is assigned the NULL tag in case of missing
labels.
It is then ignored (i.e., masked) when computing the
cross-entropy loss.
To avoid predicting NULL, we set the corresponding output cell in
the softmax to $-\infty$, similarly to \citet{eskander-etal-2020-unsupervised}.

\section{Experimental setup \label{sec:protocol}}

\begin{table}[!t]
	\centering
	\small
	\resizebox{.45\textwidth}{!}{
		\begin{tabular}{llllr}
			\toprule
			&Lang    & ISO & Family  &  \# verses    \\ \midrule
			\multirow{15}{*}{\rotatebox{90}{\scriptsize\begin{tabular}{c}Training\\ 
						languages\end{tabular}}}
			&Arabic  & arb        &  Afro-Asiatic, Semitic	 &  31173 \\
			& Chinese & zho & Sino-Tibetan, Sinitic & 31157 \\
			& Danish & dan & Indo-European, Germanic	& 31173 \\
			& English & eng & Indo-European  & 31099 \\
			&Finnish & fin       & Uralic, 	Finnic &  30200  \\
			& French & fra &  Indo-European, Romance & 31173 \\
			& German & deu & Indo-European, Germanic  & 31173 \\
			& Irish 	&gle & Indo-European, Celtic & 34957 \\
			&Italian 	& ita &  Indo-European, Romance & 35377 \\
			& Polish 	& Pol & Indo-European, Slavic & 31157  \\
			& Russian & rus    & Indo-European, Slavic &  31173      \\
			&Spanish & spa   & Indo-European, Romance   &  31157      \\
			&Swedish & swe     & Indo-European, Germanic &  31157     \\
			&Tamil   & tam         & 	Dravidian, Southern Dravidian	  &  7942   \\ 
			&Urdu   & urd &   Indo-European, Indic        &  7046    \\ \midrule
			\multirow{3}{*}{\rotatebox{90}{\scriptsize\begin{tabular}{c}Dev\\ 
						languages\end{tabular}}}
			
			& Czech & ces & Indo-European, Slavic & 31157\\
			& Greek   & ell     & 	Indo-European, Greek &  31173      \\
			& Hebrew & heb & Afro-Asiatic, Semitic & 23174 \\
			& Hungarian & hun & Uralic, Ugric & 31157 \\  \midrule
			
			\multirow{17}{*}{\rotatebox{90}{\scriptsize\begin{tabular}{c}Test\\ 
						languages\end{tabular}}}
					
			& Afrikaans & afr & Indo-European, Germanic & 31157 \\
			& Amharic & amh & Afro-Asiatic, Semitic & 7942 \\
			& Basque & eus & Basque, Basque & 7958 \\
			& Belarusian & bel & Indo-European, Slavic   & 7958 \\
			& Bulgarian & bul & Indo-European, Slavic  & 31173 \\
			& Hindi   & hin     & Indo-European, Indic   &  7952   \\
			& Indonesian   & ind & Austronesian, Malayo-Sumbawan & 31157 \\
			& Lithuanian & lit & Indo-European, Baltic	& 31149 \\
			& Marathi & mar & Indo-European, Indic  & 7947 \\
			& Persian & pes & Indo-European, Iranian  & 7931 \\
			& Portuguese & pos &  Indo-European, Romance  & 31157 \\
			& Telugu & tel &  	Dravidian, South-Central Dravidian	 & 31163 \\
			& Turkish & tur & 	Altaic, Turkic	 & 31157 \\ 
			& Bambara & bam & 		Mande, Western Mande& 7958\\
			& Erzya & myv& 	Uralic, Mordvin	 & 7958\\
			& Manx & glv & 	 Indo-European, Celtic& 3994\\
			& Yoruba & yor & 		Niger-Congo, Defoid & 30819 \\ \bottomrule

		\end{tabular}
	}
	\caption{Language family and number of verses in PBC for training, dev, and test languages in our experiments.
		\label{tab:languages}}
\end{table}

\begin{table*}[!ht]
	\centering
	\resizebox{0.95\textwidth}{!}{%
		\begin{tabular}{rrrrrrrrrrrrrrrrrr}
			\toprule
			\multicolumn{13}{c}{with XLM-R} && \multicolumn{4}{c}{without XLM-R} \\ \cmidrule{6-8}  \cmidrule{16-17}
			afr & amh & eus & bul & hin & ind & lit & pes & por & tel & tur & bel & mar & & bam & myv & glv & yor \\  \cmidrule{1-13} \cmidrule{15-18} 
			87.7 & 82.4   & 70.9    &  90.1   &  81.8   & 85.3 &  85.7   &  81.8  &  89.2  &  83.8   &  80.1 & 85.9    &   87.9
			& & 65.4    &  64.4 & 63.9    &  59.9  \\ 
			\bottomrule
		\end{tabular}
	}
	\caption{Accuracy on UD v2.10 test
		for \ourmethodtwo when transferring from all 
		training source languages
		(i.e., \ourmethodtwo-All).
		See the other tables for comparison with
		prior work, which uses older versions of UD.
		\label{tab:main}}
\end{table*}

Table~\ref{tab:languages} gives our
split of languages into training (15), development
(4) and test (17) sets.
The training set contains the source languages used for the transfer, while the development set languages are used as targets for parameter tuning.
Training and test languages represent diverse language families and diverse availability. 
Note that training and dev languages are high-resource while test languages are low-resource.
For most of the test languages, there are fewer than 8000 verses available in the
Parallel Bible Corpus;\footnote{Bible versions are described in Appendix~\ref{appendix_data}.}  for Manx, fewer than 4000. 
We evaluate POS tagging performance on the Universal Dependencies (UD) \citep{UD} test sets.
As UD and PBC tokenizations differ, we further adopt the following rule:
if a PBC token corresponds to a sequence of several UD tokens,
we replace the sequence with the original word, tagged
with the tag of the UD token in the sequence that is highest
in the dependency tree
(cf.\ \cite{agic2016multilingual}).
To tag the high-resource training and dev languages, we use
Stanza
\citep{qi-etal-2020-stanza},\footnote{\url{https://stanfordnlp.github.io/stanza/}} a
state-of-the-art NLP Python library.
We create word alignments using Eflomal
\citep{Ostling2016efmaral},\footnote{\url{7github.com/robert/eflomal}} a high-quality statistical word aligner, 
with the ``intersection'' symmetrization heuristic.
Other than parallel data, Eflomal does not need any supervision signal;
we can thus use it for any language pair in PBC.
Details on models' hyperparameters are in Appendix \ref{app:model_param}.
All tagging results reported below are averages over three runs of the
neural POS tagger model.

\section{Results \label{sec:results}}

We evaluate  \ourmethod{} on 17 test languages from different families, resource availabilities, and scripts, on
Universal Dependencies v2.10, the latest version (see details in Appendix \ref{appendix_ud}).
Our results are in Table~\ref{tab:main}.
For the four languages not supported by XLM-R,
static embeddings are used 
(see \S\ref{ssec:features})
during the training in the GNN part (\ourmethodtwo), 
and XLM-R embeddings in the neural POS tagger model.\footnote{
XLM-R embeddings are used even for languages unseen during
its pretraining as they
improve performance.
This is probably due to the fact that some words (e.g.,
names) can be well represented even for an unseen language.}
The best performance, $>89$, is obtained for Bulgarian and Portuguese.
All scores with XLM-R are above 80, except for Basque. This is probably 
because no language from the same family appears in the training set.
Similarly, Turkish has the lowest performance among the other test languages.
Scores without XLM-R are overall lower, yet competitive, showing that our projection
method also works for very low-resource languages.
Prior work has used older versions of UD . We now compare
against each baseline, evaluating on the relevant version
of UD in each case.

\subsection{Annotation projection-based baselines}
In this section, we compare  with
the unsupervised SOTA in cross-lingual POS tagging via annotation projection:
ESKANDER \citep{eskander-etal-2020-unsupervised}, AGIC \cite{agic2016multilingual}
 and BUYS \cite{buys2016cross} as well as EFLOMAL.
We also compare with a semi-supervised SOTA
method that uses rapid annotation in addition to cross-lingual projection: CTRL \cite{cotterell-heigold-2017-cross}. 

\subsubsection{Fully unsupervised baselines}

\begin{table*}[!ht]
	\centering
		\resizebox{0.95\textwidth}{!}{%
	\begin{tabular}{lrrrrrrrrrrrrrrrr}
		\toprule
								&  afr    & amh & eus      & bul     & hin   & ind & lit  & pes    & por  & tel & tur   &AVG & & bel    & mar  &AVG   \\  \cmidrule{2-13} \cmidrule{15-17}
		EFLOMAL-Eng 			& 73.7    & 74.9  &60.4       & 78.9   & 58.1   & 72.4 & 80.3     & 59.2     &  74.1  & 77.5 & 67.6 & 70.6
		&& 76.2 &  73.2 & 71.3 \\
		
		EFLOMAL-All 			& 83.9    & 79.3  & 64.5 & 85.0 & 68.1 & 78.4  & 82.8 & 68.6   & 83.8 & 77.1 & 74.8 & 76.9
		&& 79.6 & 77.8  & 77.2\\
		
		ESKANDER-Eng     			&  86.9   & 75.3  &   67.3 & 85.6 & 73.9       &   \textbf{84.1}   &  80.9  &  77.2 &    86.1  &   80.0 &   74.3  & 79.2
		&&  &  &  \\
		ESKANDER-All   			& 89.3    &  79.3 & 67.1   & 88.2  & 72.8  & 83.0 & 82.5  & 77.3 & 87.8 & 77.1 & 74.6 & 79.9
		& &  &  & \\ 
		
		\midrule
		
		\ourmethodone-Eng     &  86.6  & 81.9  & 67.5    &  85.7 &  76.8  & 82.7   &  81.1    &   76.2    & 87.6     &  82.5 &  76.4   & 80.4
		&  &  80.0 &  82.3 & 80.6 \\
		
		\ourmethodtwo-Eng     &  84.4  & 81.9  &   68.6   &   84.0  & 75.8  & 81.3 & 81.0 &  73.5    &  86.4   & 80.6    & 75.8   & 79.4
		& & 75.1 &   81.5  & 79.2 \\ 
		
		\ourmethodone-All     &\textbf{89.7}&  \textbf{83.6}  & 67.4    &  \textbf{89.7}   & 79.9 &   82.8     &  \textbf{85.9}      &   79.6     &  87.7  &   81.4     & \textbf{80.3}  & 82.5
		  &&    87.9  & 83.2   &  83.0 \\
		\ourmethodtwo-All     &   87.5    & 82.9  &   \textbf{70.6}  & \textbf{89.7} & \textbf{81.9}   &  83.4  &  85.8   & \textbf{81.9}  & \textbf{89.6}   &  \textbf{83.7}   & 78.4 & \textbf{83.2}
		 &&    \textbf{88.8}    &   \textbf{88.4}  & \textbf{84.0} \\ \bottomrule
		
	\end{tabular}
}
	\caption{Accuracy on UD v2.5 test  for EFLOMAL, ESKANDER \citep{eskander-etal-2020-unsupervised}
		and \ourmethod{}.
                ``-Eng'': transfer from English
                only. ``-All'': transfer from all training
                languages
                (see \citet{eskander-etal-2020-unsupervised} and Table~\ref{tab:languages}).
		Bold:  best score for each language. 
		\label{tab:comparison}
	}
\end{table*}

EFLOMAL is a simple projection method using alignment links
followed by majority voting, similar to early 
annotation projection methods
\cite{agic2015if, fossum-abney-2005-automatically}.
We first align all target sentences with the corresponding sentences in all training languages
with Eflomal \citep{Ostling2016efmaral}. 
Each target word is then tagged with the most 
common tag in the aligned source words.
The annotation projection method
ESKANDER \citep{eskander-etal-2020-unsupervised} 
uses alignment links and token and type 
constraints to project tags from source to target.
The neural POS tagger features include
XLM-R embeddings, affix embeddings, and word clusters created on PBC and Wikipedia 
of the target languages. 
Table~\ref{tab:comparison} compares 
EFLOMAL, ESKANDER and \ourmethod.
In this table -Eng stands for when 
only English is used as the source language in \ourmethod{} 
and -All stands for 
when all training languages are used (see \S\ref{paragarph:engall}).
\ourmethod  outperforms both baselines in all cases but Indonesian, where 
ESKANDER is 0.7 points better. However, they tune their hyperparameters on this language using dev data while we only tune them on 
dev languages. Compared to ESKANDER, we use a simpler neural
POS tagger and less resources, as we do not use affix embeddings nor word clusters.
Our initial experiments indicated that word clusters were not helping
in our setup. The higher quality 
of the annotated data created by \ourmethod{} may already
contain the information provided by word clusters.

\begin{table}[!ht]
	
	\centering
	\resizebox{0.4\textwidth}{!}{%
		\begin{tabular}{lllllrr}
			\toprule
			& Target && \multicolumn{2}{c}{AGIC}           && \ourmethodtwo-All            \\  \cmidrule{2-2} \cmidrule{4-5} \cmidrule{7-7} 
			\multirow{5}{*}{\rotatebox{90}{\small\begin{tabular}{c}v1.2
			\end{tabular}}}
			& bul    && 70.0 &mul  && \textbf{86.1} \\
			& hin    &&  50.5 &mul &&  \textbf{79.0} \\
			& ind    && 75.5 &mul && \textbf{79.5} \\
			& pes   && 33.7 &mul   &&  \textbf{75.2} \\
			& por   &&  84.2 &mul &&  \textbf{87.7}     \\ \midrule
			
			& Target && \multicolumn{2}{c}{BUYS}                    && \ourmethodtwo-All     \\  \cmidrule{2-2} \cmidrule{4-5} \cmidrule{7-7} 
			\multirow{2}{*}{\rotatebox{90}{\small\begin{tabular}{c}v1.2
			\end{tabular}}}
			& bul    &&   81.8 & eng     && \textbf{86.1}    \\ 
			& por    &&   \textbf{88.0}  &esp    &&  87.7      \\ \midrule

			& Target && \multicolumn{2}{c}{CTRL}    &&\ourmethodtwo-All     \\  \cmidrule{2-2} \cmidrule{4-5} \cmidrule{7-7}
			\multirow{4}{*}{\rotatebox{90}{\small\begin{tabular}{c}
						v2.0\end{tabular}}}
			& Bul   &&   68.8 &rus-100    &&    \textbf{89.3}           \\
			& Bul   &&   83.1 &rus-1000     &&    \textbf{89.3}     \\
			& Por    &&   81.8 & esp-100 &&     \textbf{90.1}  \\ 
			& Por   &&   88.9 &esp-1000  &&   \textbf{90.1}   \\
			\bottomrule
		\end{tabular}
	}
	\caption{Accuracy on UD test for AGIC \citep{agic2016multilingual},
		BUYS \citep{buys2016cross},
		CTRL \citep{cotterell-heigold-2017-cross}
		and  \ourmethodtwo. We also report the source language or ``mul'' for multilingual, and for CTRL, the number of the supervision tokens.
		\label{tab:sota}}
\end{table}
Table~\ref{tab:sota} compares AGIC, BUYS, CTRL, and \ourmethodtwo{}.
AGIC \cite{agic2016multilingual} is a cross-lingual POS
tagger for low-resource languages
based on PBC excerpts
and translations of the Watchtower.\footnote{Obtained by crawling \url{http://wol.jw.org}}
BUYS \citep{buys2016cross} extends previous approaches for projecting POS tags using bitexts
to infer constraints on the possible tags for a given word type or token.

Table~\ref{tab:sota} shows that \ourmethod
outperforms AGIC and BUYS, except for Portuguese (BUYS),
where our results are slightly below. BUYS projects from Spanish, 
which is closely related to Portuguese.
\citet{eskander-etal-2020-unsupervised} showed that it can be advantageous to transfer only from one closely
related language as opposed to a mix of close and distant languages.
Note that BUYS performance for Portuguese drops down to 84.3 when
transferring from English.
BUYS also uses Europarl\footnote{\url{http://www.statmt.org/europarl/
}} with up to 2M tokens which is closer
in domain to UD than PBC.
Thus, compared to BUYS, the parallel data we use are smaller, and from a more distant domain.

\subsubsection{Semisupervised baseline}
CTRL \citep{cotterell-heigold-2017-cross} is a character-level recurrent neural network
for multi-task cross-lingual transfer of morphological
taggers. Their experiments include small sets of 100 and 1000 annotated target tokens. 
The bottom part of Table~\ref{tab:sota} shows that
\ourmethodtwo outperforms CTRL despite being fully unsupervised.

\subsection{Zero-shot baselines}

Cross-lingual projection is also possible thanks
to multilingual pretrained language models (PLMs). 
A PLM is first fine-tuned to POS tagging on source
languages and then used to infer tags for target
languages. 
While this approach performs well for some
languages without requiring any parallel data, its
performance tends to be poor for low-resource languages \citep{hu-etal-2021-explicit}.
\citet{joshi-etal-2020-state} cluster languages into six groups based on the amount of 
available unlabeled and labeled data that exists for them. Groups 1 and 2 consist of
languages such as Manx and Yoruba with the least amount of available data, while group
5 contains languages like  English and Spanish with the largest amount of available monolingual and labeled data. 
We compare our approach with three baselines using test languages from groups 1 and 2. 

\paragraph{mBERT based baselines:}
\citet{kondratyuk-straka-2019-75} use the zero-shot approach
with multilingual BERT \citep{devlin-etal-2019-bert} as PLM. 
We train our POS taggers using mBERT (instead of XLM-R) embeddings for a fair comparison.
Table \ref{tab:zero_shot} displays the results for the low-resource languages in group 1 and 2, 
%of 
which are also reported in the compared work.
\ourmethodtwo outperforms zero-shot  in all cases by at least
12 percentage points. This result suggests that annotation projection using 
\ourmethod{} is more effective than using multilingual representations for truly low-resource languages
(i.e., languages from the first two groups of \citet{joshi-etal-2020-state}). 
To create proper representations for a language, PLMs require a huge amount of 
monolingual data that is not available for many languages. As Table~\ref{tab:zero_shot}
suggests, due to poor representations, zero-shot transfer to these languages is also poor.
However, we were able to successfully
exploit the Bible's parallel data in \ourmethod{} for the benefit of these languages.

\begin{table}[!th]
	
	\centering
		\resizebox{0.45\textwidth}{!}{%
	\begin{tabular}{llrrrrrr}
		\toprule
		        & bam && myv    && yor       \\  \cmidrule{2-2}\cmidrule{4-4}\cmidrule{6-6}
		\citet{kondratyuk-straka-2019-75} & 30.9 && 46.7 && 50.9 \\
		\ourmethodtwo-ALL  & \textbf{65.5} &&  \textbf{64.6} && \textbf{63.3}  \\ \bottomrule
	\end{tabular}
}
	\caption{POS tagging accuracy on UD v2.3 test for zero-shot
		mBERT and \ourmethodtwo using mBERT
		embeddings.\label{tab:zero_shot}}
\end{table}

\paragraph{XLM-R based baselines:}
\label{xlmrbasedbaselines}
\citet{ebrahimi-kann-2021-adapt} continue pretraining PLMs on PBC  
and show that
this boosts performance for languages unseen during the initial pretraining.
\citet{wang-etal-2022-expanding} adapt PLMs to languages with little 
monolingual data using various sources of data including PanLex lexicons,\footnote{\url{https://panlex.org/snapshot/}} 
translations of English Wikipedia to target languages
and the JHU Bible corpus \citep{mccarthy-etal-2020-johns}.
These approaches are in fact complementary to \ourmethod{}:
we can equip \ourmethod{} with better multilingual representations to further
improve our results based on standard XLM-R.
This is reflected in Table~\ref{tab:wang}, where we report results for
zero-shot baselines and combinations 
based on \citet{wang-etal-2022-expanding}'s improved XLM-R
embeddings (instead of standard XLM-R) to represent tokens for the POS tagger.
We see that these combinations lead to large performance improvements,
establishing new SOTA results.

\begin{table}[!t]
	\centering
		\resizebox{0.45\textwidth}{!}{%
	\begin{tabular}{lrrrrr}
		\toprule
		& bam && myv && glv \\ \cmidrule{2-2}\cmidrule{4-4}\cmidrule{6-6}
		\citet{ebrahimi-kann-2021-adapt} & 60.5 && 66.6 && 59.7 \\
		\citet{wang-etal-2022-expanding} & 69.4 &&  74.3    &&  68.8 \\ 
		\cmidrule{1-6}
		 \ourmethodtwo-ALL + wang-before & \textbf{71.1} &&78.9  && 70.1 \\
		 \ourmethodtwo-ALL + wang-after & 70.2 && \textbf{80.6} && \textbf{70.7} \\
		\bottomrule
	\end{tabular}
}
	\caption{Accuracy on UD v2.5 test for two baselines and for our method combined with \citep{wang-etal-2022-expanding}'s XLM-R models 
          before and after finetuning on the POS tagging task. (``glv'' accuracy is on v2.7.)
		\label{tab:wang}}
\end{table}

\section{Analysis \label{sec:analysis}}

\subsection{Ablation study}

We conduct an ablation study to better understand what benefits our model. 

\paragraph{``Eng'' vs ``All''} \label{paragarph:engall}
Previous works highlighted the importance of a diverse set of source languages for cross-lingual transfer \citep{lin-etal-2019-choosing,turc2021revisiting}.
The last four lines of Table~\ref{tab:comparison} report
\ourmethodone and \ourmethodtwo results when
transferring from English (i.e., using English as the only
source),
and when transferring from the full
set of source languages (see Table \ref{tab:languages}). 
The transfer from English
has lower performance than from all languages (except for
a decrease from 67.5 to 67.4 for Basque/\ourmethodone). This means
that our projection method does benefit from more data and
from the rich information present in the diversity of source
languages.

\paragraph{\ourmethodone vs \ourmethodtwo }

Table~\ref{tab:comparison} reports results when training the neural POS tagger on \ourmethodone data and on \ourmethodtwo data. 
\ourmethodone performs better than \ourmethodtwo for four languages: Afrikaans, Lithuanian, Portuguese, and Turkish; but the performance difference is small (1.2 percentage points difference on average).
In eight out of thirteen languages, \ourmethodtwo gives better results (2.3 percentage points difference on average).
This shows that the transformer architecture and the self-learning strategy are effective for most languages. 

\paragraph{Contextualized vs.\ Static embeddings}
Our \ourmethod{} models use XLM-R embeddings for languages for which they
are available, otherwise static embeddings
(see \S\ref{ssec:features}).
In order to understand their usefulness in the transfer
process, we compare with the performance obtained when
static embeddings are 
used by \ourmethodtwo.
Results reported in Appendix \ref{appendix_results} show an average improvement of 3
percentage points when XLM-R embeddings are used.
The largest differences ($>5$\%) are observed for Hindi, Persian, and Marathi.
However, for the four languages not supported by XLM-R, the
POS tagging accuracy
is substantially lower when using
contextualized embeddings compared to static embeddings (16.6
points drop on average).

\subsection{Quality of artificial training sets}
\begin{figure}[!t] \centering 
	\includegraphics[width=1\linewidth, ]{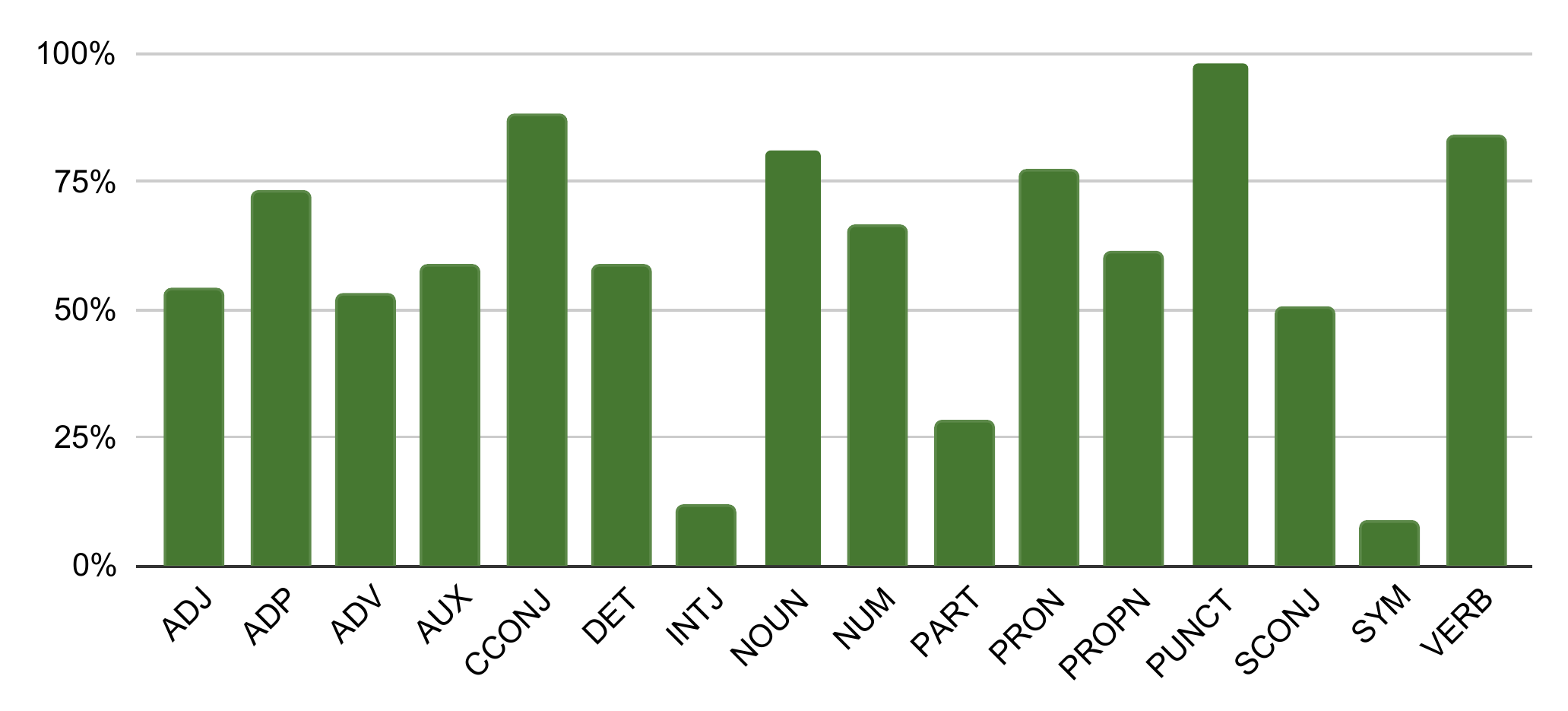} 
	\caption{Average per tag accuracy of our \ourmethod sets with respect to the ``silver'' reference.}
	\label{fig:appendix_plot}
\end{figure}
In order to evaluate the quality of the training sets
generated by \ourmethodtwo
(``\ourmethod sets''), we create a
``silver'' reference and compute the accuracy of \ourmethod
sets with respect to it. To build the silver reference, we
annotate the training sets with the Stanza POS tagger for
the languages for which it is available (12 out of 17). We obtain
an average accuracy of 78.7, with Belarusian being the best and
Basque the worst. 
The best predicted tokens are punctuation marks,
coordinating conjunctions, and verbs, while the worst ones are
symbols,  interjections, and particles (see Figure \ref{fig:appendix_plot}).
The high accuracy of 78.7 illustrates
the ability of \ourmethodtwo to successfully project annotations from
high to low-resource languages.

\section{Conclusion and future work}
We presented \ourmethod{}, a
novel method for transferring labels from high-resource
source to low-resource target languages, based on a formalization
of annotation projection
as graph-based label
propagation.  We exploited the Parallel Bible Corpus
and showed that reasonably accurate POS taggers can be
bootstrapped from projected labels.  Since we do not
use PBC-specific or language-specific features, \ourmethod{} is in
principle applicable to the more than 1000 languages of PBC and to
any other multiparallel corpus.

One direction for the future is to employ a similar model to transfer higher-level 
structures such as dependency trees. Since our method works with 
graph structures, one might be able to project dependency trees effectively.
We could also extend our projection method to other tagging
tasks like named entity recognition -- although this requires
using other parallel corpora to mitigate the domain shift
problem of such a task.
Another line for future work is to study the best combinations of source languages to transfer to any target language.

\section*{Limitations} % It will not count towards the page limit.  
Our method is evaluated on 17 languages carefully chosen to be from different families 
and scripts. However, we don't consider the other  languages
(more than 1000) in PBC due to 
computational constraints and lack of test sets.

A limitation of the \ourmethod{} is that training over a MAG (multilingual alignment graph)
created for all  PBC languages requires a prohibitively large amount of resources,
and based on our experiments, if we use a larger number of target languages at the same time,
the performance will likely drop. 
Therefore one has to process languages in smaller batches (in our case, $36$ languages).
Accordingly, to cover all PBC subcorpora, $1341/36 = 38$ \ourmethod{} models should in
principle be trained.

\section*{Ethic statement} 

Our work is based on the Parallel Bible Corpus of \citet{mayer-cysouw-2014-creating}
that consists of  Bible verses and  is tested on the Universal Dependency treebanks
\citep{UD}, an ensemble of different data sources.
We would like to clarify that we treat the data simply as a multiparallel corpus, and the
content does not necessarily reflect the opinions of
the authors nor of the institutions funding the authors.

\section*{Acknowledgements}
We would like to thank Sebastian Ruder for his kind and thoughtful suggestions on this work.

This work was funded by the European Research Council (grant \#740516) and the German Federal Ministry of Education and Research (BMBF, grant \#01IS18036A). 

\bibliography{anthology,custom}

\begin{thebibliography}{62}
\expandafter\ifx\csname natexlab\endcsname\relax\def\natexlab#1{#1}\fi

\bibitem[{Agi{\'c} et~al.(2015{\natexlab{a}})Agi{\'c}, Hovy, and
  S{\o}gaard}]{agic-etal-2015-bit}
{\v{Z}}eljko Agi{\'c}, Dirk Hovy, and Anders S{\o}gaard. 2015{\natexlab{a}}.
\newblock \href {https://doi.org/10.3115/v1/P15-2044} {If all you have is a bit
  of the {B}ible: Learning {POS} taggers for truly low-resource languages}.
\newblock In \emph{Proceedings of the 53rd Annual Meeting of the Association
  for Computational Linguistics and the 7th International Joint Conference on
  Natural Language Processing (Volume 2: Short Papers)}, pages 268--272,
  Beijing, China. Association for Computational Linguistics.

\bibitem[{Agi{\'c} et~al.(2015{\natexlab{b}})Agi{\'c}, Hovy, and
  S{\o}gaard}]{agic2015if}
{\v{Z}}eljko Agi{\'c}, Dirk Hovy, and Anders S{\o}gaard. 2015{\natexlab{b}}.
\newblock If all you have is a bit of the bible: Learning pos taggers for truly
  low-resource languages.
\newblock In \emph{Proceedings of the 53rd Annual Meeting of the Association
  for Computational Linguistics and the 7th International Joint Conference on
  Natural Language Processing (Volume 2: Short Papers)}, pages 268--272.

\bibitem[{Agi{\'c} et~al.(2016)Agi{\'c}, Johannsen, Plank, Alonso, Schluter,
  and S{\o}gaard}]{agic2016multilingual}
{\v{Z}}eljko Agi{\'c}, Anders Johannsen, Barbara Plank,
  H{\'e}ctor~Mart{\'\i}nez Alonso, Natalie Schluter, and Anders S{\o}gaard.
  2016.
\newblock Multilingual projection for parsing truly low-resource languages.
\newblock \emph{Transactions of the Association for Computational Linguistics},
  4:301--312.

\bibitem[{Agi{\'c} and Vuli{\'c}(2019)}]{agic-vulic-2019-jw300}
{\v{Z}}eljko Agi{\'c} and Ivan Vuli{\'c}. 2019.
\newblock \href {https://doi.org/10.18653/v1/P19-1310} {{JW}300: A
  wide-coverage parallel corpus for low-resource languages}.
\newblock In \emph{Proceedings of the 57th Annual Meeting of the Association
  for Computational Linguistics}, pages 3204--3210, Florence, Italy.
  Association for Computational Linguistics.

\bibitem[{Akbik et~al.(2019)Akbik, Bergmann, Blythe, Rasul, Schweter, and
  Vollgraf}]{akbik-etal-2019-flair}
Alan Akbik, Tanja Bergmann, Duncan Blythe, Kashif Rasul, Stefan Schweter, and
  Roland Vollgraf. 2019.
\newblock \href {https://doi.org/10.18653/v1/N19-4010} {{FLAIR}: An easy-to-use
  framework for state-of-the-art {NLP}}.
\newblock In \emph{Proceedings of the 2019 Conference of the North {A}merican
  Chapter of the Association for Computational Linguistics (Demonstrations)},
  pages 54--59, Minneapolis, Minnesota. Association for Computational
  Linguistics.

\bibitem[{Bastings et~al.(2017)Bastings, Titov, Aziz, Marcheggiani, and
  Sima{'}an}]{bastings-etal-2017-graph}
Jasmijn Bastings, Ivan Titov, Wilker Aziz, Diego Marcheggiani, and Khalil
  Sima{'}an. 2017.
\newblock \href {https://doi.org/10.18653/v1/D17-1209} {Graph convolutional
  encoders for syntax-aware neural machine translation}.
\newblock In \emph{Proceedings of the 2017 Conference on Empirical Methods in
  Natural Language Processing}, pages 1957--1967, Copenhagen, Denmark.
  Association for Computational Linguistics.

\bibitem[{Beck et~al.(2018)Beck, Haffari, and Cohn}]{beck-etal-2018-graph}
Daniel Beck, Gholamreza Haffari, and Trevor Cohn. 2018.
\newblock \href {https://doi.org/10.18653/v1/P18-1026} {Graph-to-sequence
  learning using gated graph neural networks}.
\newblock In \emph{Proceedings of the 56th Annual Meeting of the Association
  for Computational Linguistics (Volume 1: Long Papers)}, pages 273--283,
  Melbourne, Australia. Association for Computational Linguistics.

\bibitem[{Boldi and Vigna(2014)}]{boldi2014axioms}
Paolo Boldi and Sebastiano Vigna. 2014.
\newblock Axioms for centrality.
\newblock \emph{Internet Mathematics}, 10(3-4):222--262.

\bibitem[{Brandes(2001)}]{brandes2001faster}
Ulrik Brandes. 2001.
\newblock A faster algorithm for betweenness centrality.
\newblock \emph{Journal of mathematical sociology}, 25(2):163--177.

\bibitem[{Brown et~al.(1992)Brown, Della~Pietra, deSouza, Lai, and
  Mercer}]{brown-etal-1992-class}
Peter~F. Brown, Vincent~J. Della~Pietra, Peter~V. deSouza, Jenifer~C. Lai, and
  Robert~L. Mercer. 1992.
\newblock \href {https://aclanthology.org/J92-4003} {Class-based
  \textit{n}-gram models of natural language}.
\newblock \emph{Computational Linguistics}, 18(4):467--480.

\bibitem[{Buys and Botha(2016{\natexlab{a}})}]{buys-botha-2016-cross}
Jan Buys and Jan~A. Botha. 2016{\natexlab{a}}.
\newblock \href {https://doi.org/10.18653/v1/P16-1184} {Cross-lingual
  morphological tagging for low-resource languages}.
\newblock In \emph{Proceedings of the 54th Annual Meeting of the Association
  for Computational Linguistics (Volume 1: Long Papers)}, pages 1954--1964,
  Berlin, Germany. Association for Computational Linguistics.

\bibitem[{Buys and Botha(2016{\natexlab{b}})}]{buys2016cross}
Jan Buys and Jan~A Botha. 2016{\natexlab{b}}.
\newblock Cross-lingual morphological tagging for low-resource languages.
\newblock \emph{arXiv preprint arXiv:1606.04279}.

\bibitem[{Clauset et~al.(2004)Clauset, Newman, and Moore}]{clauset2004finding}
Aaron Clauset, Mark~EJ Newman, and Cristopher Moore. 2004.
\newblock Finding community structure in very large networks.
\newblock \emph{Physical review E}, 70(6):066111.

\bibitem[{Conneau et~al.(2020)Conneau, Khandelwal, Goyal, Chaudhary, Wenzek,
  Guzm{\'a}n, Grave, Ott, Zettlemoyer, and
  Stoyanov}]{conneau-etal-2020-unsupervised}
Alexis Conneau, Kartikay Khandelwal, Naman Goyal, Vishrav Chaudhary, Guillaume
  Wenzek, Francisco Guzm{\'a}n, Edouard Grave, Myle Ott, Luke Zettlemoyer, and
  Veselin Stoyanov. 2020.
\newblock \href {https://doi.org/10.18653/v1/2020.acl-main.747} {Unsupervised
  cross-lingual representation learning at scale}.
\newblock In \emph{Proceedings of the 58th Annual Meeting of the Association
  for Computational Linguistics}, pages 8440--8451, Online. Association for
  Computational Linguistics.

\bibitem[{Cordasco and Gargano(2010)}]{cordasco2010community}
Gennaro Cordasco and Luisa Gargano. 2010.
\newblock Community detection via semi-synchronous label propagation
  algorithms.
\newblock In \emph{2010 IEEE international workshop on: business applications
  of social network analysis (BASNA)}, pages 1--8. IEEE.

\bibitem[{Cotterell and Heigold(2017)}]{cotterell-heigold-2017-cross}
Ryan Cotterell and Georg Heigold. 2017.
\newblock \href {https://doi.org/10.18653/v1/D17-1078} {Cross-lingual
  character-level neural morphological tagging}.
\newblock In \emph{Proceedings of the 2017 Conference on Empirical Methods in
  Natural Language Processing}, pages 748--759, Copenhagen, Denmark.
  Association for Computational Linguistics.

\bibitem[{Devlin et~al.(2019)Devlin, Chang, Lee, and
  Toutanova}]{devlin-etal-2019-bert}
Jacob Devlin, Ming-Wei Chang, Kenton Lee, and Kristina Toutanova. 2019.
\newblock \href {https://doi.org/10.18653/v1/N19-1423} {{BERT}: Pre-training of
  deep bidirectional transformers for language understanding}.
\newblock In \emph{Proceedings of the 2019 Conference of the North {A}merican
  Chapter of the Association for Computational Linguistics: Human Language
  Technologies, Volume 1 (Long and Short Papers)}, pages 4171--4186,
  Minneapolis, Minnesota. Association for Computational Linguistics.

\bibitem[{Dufter et~al.(2018)Dufter, Zhao, Schmitt, Fraser, and
  Sch{\"u}tze}]{dufter2018embedding}
Philipp Dufter, Mengjie Zhao, Martin Schmitt, Alexander Fraser, and Hinrich
  Sch{\"u}tze. 2018.
\newblock Embedding learning through multilingual concept induction.
\newblock \emph{arXiv preprint arXiv:1801.06807}.

\bibitem[{Duong et~al.(2019)Duong, Hoang, Dang, Nguyen, and
  Aberer}]{duong2019node}
Chi~Thang Duong, Thanh~Dat Hoang, Ha~The~Hien Dang, Quoc Viet~Hung Nguyen, and
  Karl Aberer. 2019.
\newblock On node features for graph neural networks.
\newblock \emph{arXiv preprint arXiv:1911.08795}.

\bibitem[{Duvenaud et~al.(2015)Duvenaud, Maclaurin, Iparraguirre, Bombarell,
  Hirzel, Aspuru-Guzik, and Adams}]{duvenaud2015convolutional}
David~K Duvenaud, Dougal Maclaurin, Jorge Iparraguirre, Rafael Bombarell,
  Timothy Hirzel, Al{\'a}n Aspuru-Guzik, and Ryan~P Adams. 2015.
\newblock Convolutional networks on graphs for learning molecular fingerprints.
\newblock \emph{Advances in neural information processing systems}, 28.

\bibitem[{Ebrahimi and Kann(2021)}]{ebrahimi-kann-2021-adapt}
Abteen Ebrahimi and Katharina Kann. 2021.
\newblock \href {https://doi.org/10.18653/v1/2021.acl-long.351} {How to adapt
  your pretrained multilingual model to 1600 languages}.
\newblock In \emph{Proceedings of the 59th Annual Meeting of the Association
  for Computational Linguistics and the 11th International Joint Conference on
  Natural Language Processing (Volume 1: Long Papers)}, pages 4555--4567,
  Online. Association for Computational Linguistics.

\bibitem[{Eskander et~al.(2020)Eskander, Muresan, and
  Collins}]{eskander-etal-2020-unsupervised}
Ramy Eskander, Smaranda Muresan, and Michael Collins. 2020.
\newblock \href {https://doi.org/10.18653/v1/2020.emnlp-main.391} {Unsupervised
  cross-lingual part-of-speech tagging for truly low-resource scenarios}.
\newblock In \emph{Proceedings of the 2020 Conference on Empirical Methods in
  Natural Language Processing (EMNLP)}, pages 4820--4831, Online. Association
  for Computational Linguistics.

\bibitem[{Fang and Cohn(2016)}]{fang-cohn-2016-learning}
Meng Fang and Trevor Cohn. 2016.
\newblock \href {https://doi.org/10.18653/v1/K16-1018} {Learning when to trust
  distant supervision: An application to low-resource {POS} tagging using
  cross-lingual projection}.
\newblock In \emph{Proceedings of The 20th {SIGNLL} Conference on Computational
  Natural Language Learning}, pages 178--186, Berlin, Germany. Association for
  Computational Linguistics.

\bibitem[{Fossum and Abney(2005)}]{fossum-abney-2005-automatically}
Victoria Fossum and Steven Abney. 2005.
\newblock \href {https://doi.org/10.1007/11562214_75} {Automatically inducing a
  part-of-speech tagger by projecting from multiple source languages across
  aligned corpora}.
\newblock In \emph{Second International Joint Conference on Natural Language
  Processing: Full Papers}.

\bibitem[{Freeman(1978)}]{freeman1978centrality}
Linton~C Freeman. 1978.
\newblock Centrality in social networks conceptual clarification.
\newblock \emph{Social networks}, 1(3):215--239.

\bibitem[{Ganchev and Das(2013)}]{ganchev-das-2013-cross}
Kuzman Ganchev and Dipanjan Das. 2013.
\newblock \href {https://aclanthology.org/D13-1205} {Cross-lingual
  discriminative learning of sequence models with posterior regularization}.
\newblock In \emph{Proceedings of the 2013 Conference on Empirical Methods in
  Natural Language Processing}, pages 1996--2006, Seattle, Washington, USA.
  Association for Computational Linguistics.

\bibitem[{Garrette and Baldridge(2013)}]{garrette-baldridge-2013-learning}
Dan Garrette and Jason Baldridge. 2013.
\newblock \href {https://aclanthology.org/N13-1014} {Learning a part-of-speech
  tagger from two hours of annotation}.
\newblock In \emph{Proceedings of the 2013 Conference of the North {A}merican
  Chapter of the Association for Computational Linguistics: Human Language
  Technologies}, pages 138--147, Atlanta, Georgia. Association for
  Computational Linguistics.

\bibitem[{Hochreiter and Schmidhuber(1997)}]{hochreiter1997long}
Sepp Hochreiter and J{\"u}rgen Schmidhuber. 1997.
\newblock Long short-term memory.
\newblock \emph{Neural computation}, 9(8):1735--1780.

\bibitem[{Hu et~al.(2021)Hu, Johnson, Firat, Siddhant, and
  Neubig}]{hu-etal-2021-explicit}
Junjie Hu, Melvin Johnson, Orhan Firat, Aditya Siddhant, and Graham Neubig.
  2021.
\newblock \href {https://doi.org/10.18653/v1/2021.naacl-main.284} {Explicit
  alignment objectives for multilingual bidirectional encoders}.
\newblock In \emph{Proceedings of the 2021 Conference of the North American
  Chapter of the Association for Computational Linguistics: Human Language
  Technologies}, pages 3633--3643, Online. Association for Computational
  Linguistics.

\bibitem[{Imani et~al.(2022)Imani, Senel, Jalili~Sabet, Yvon, and
  Schuetze}]{imani-etal-2022-graph}
Ayyoob Imani, L{\"u}tfi~Kerem Senel, Masoud Jalili~Sabet, Fran{\c{c}}ois Yvon,
  and Hinrich Schuetze. 2022.
\newblock \href {https://aclanthology.org/2022.findings-acl.108} {Graph neural
  networks for multiparallel word alignment}.
\newblock In \emph{Findings of the Association for Computational Linguistics:
  ACL 2022}, pages 1384--1396, Dublin, Ireland. Association for Computational
  Linguistics.

\bibitem[{ImaniGooghari et~al.(2021)ImaniGooghari, Jalili~Sabet, Dufter, Cysou,
  and Sch{\"u}tze}]{imanigooghari-etal-2021-parcoure}
Ayyoob ImaniGooghari, Masoud Jalili~Sabet, Philipp Dufter, Michael Cysou, and
  Hinrich Sch{\"u}tze. 2021.
\newblock \href {https://doi.org/10.18653/v1/2021.acl-demo.8} {{P}ar{C}our{E}:
  A parallel corpus explorer for a massively multilingual corpus}.
\newblock In \emph{Proceedings of the 59th Annual Meeting of the Association
  for Computational Linguistics and the 11th International Joint Conference on
  Natural Language Processing: System Demonstrations}, pages 63--72, Online.
  Association for Computational Linguistics.

\bibitem[{Joshi et~al.(2020)Joshi, Santy, Budhiraja, Bali, and
  Choudhury}]{joshi-etal-2020-state}
Pratik Joshi, Sebastin Santy, Amar Budhiraja, Kalika Bali, and Monojit
  Choudhury. 2020.
\newblock \href {https://doi.org/10.18653/v1/2020.acl-main.560} {The state and
  fate of linguistic diversity and inclusion in the {NLP} world}.
\newblock In \emph{Proceedings of the 58th Annual Meeting of the Association
  for Computational Linguistics}, pages 6282--6293, Online. Association for
  Computational Linguistics.

\bibitem[{Kondratyuk and Straka(2019)}]{kondratyuk-straka-2019-75}
Dan Kondratyuk and Milan Straka. 2019.
\newblock \href {https://doi.org/10.18653/v1/D19-1279} {75 languages, 1 model:
  Parsing {U}niversal {D}ependencies universally}.
\newblock In \emph{Proceedings of the 2019 Conference on Empirical Methods in
  Natural Language Processing and the 9th International Joint Conference on
  Natural Language Processing (EMNLP-IJCNLP)}, pages 2779--2795, Hong Kong,
  China. Association for Computational Linguistics.

\bibitem[{Levy et~al.(2017)Levy, S{\o}gaard, and
  Goldberg}]{levy-etal-2017-strong}
Omer Levy, Anders S{\o}gaard, and Yoav Goldberg. 2017.
\newblock \href {https://aclanthology.org/E17-1072} {A strong baseline for
  learning cross-lingual word embeddings from sentence alignments}.
\newblock In \emph{Proceedings of the 15th Conference of the {E}uropean Chapter
  of the Association for Computational Linguistics: Volume 1, Long Papers},
  pages 765--774, Valencia, Spain. Association for Computational Linguistics.

\bibitem[{Li et~al.(2020)Li, Chen, Zhao, Zhang, Wang, and Tian}]{li2020dynamic}
Maosen Li, Siheng Chen, Yangheng Zhao, Ya~Zhang, Yanfeng Wang, and Qi~Tian.
  2020.
\newblock Dynamic multiscale graph neural networks for 3d skeleton based human
  motion prediction.
\newblock In \emph{Proceedings of the IEEE/CVF Conference on Computer Vision
  and Pattern Recognition}, pages 214--223.

\bibitem[{Li et~al.(2012)Li, Gra{\c{c}}a, and Taskar}]{li-etal-2012-wiki}
Shen Li, Jo{\~a}o Gra{\c{c}}a, and Ben Taskar. 2012.
\newblock \href {https://aclanthology.org/D12-1127} {{W}iki-ly supervised
  part-of-speech tagging}.
\newblock In \emph{Proceedings of the 2012 Joint Conference on Empirical
  Methods in Natural Language Processing and Computational Natural Language
  Learning}, pages 1389--1398, Jeju Island, Korea. Association for
  Computational Linguistics.

\bibitem[{Lin et~al.(2019)Lin, Chen, Lee, Li, Zhang, Xia, Rijhwani, He, Zhang,
  Ma, Anastasopoulos, Littell, and Neubig}]{lin-etal-2019-choosing}
Yu-Hsiang Lin, Chian-Yu Chen, Jean Lee, Zirui Li, Yuyan Zhang, Mengzhou Xia,
  Shruti Rijhwani, Junxian He, Zhisong Zhang, Xuezhe Ma, Antonios
  Anastasopoulos, Patrick Littell, and Graham Neubig. 2019.
\newblock \href {https://doi.org/10.18653/v1/P19-1301} {Choosing transfer
  languages for cross-lingual learning}.
\newblock In \emph{Proceedings of the 57th Annual Meeting of the Association
  for Computational Linguistics}, pages 3125--3135, Florence, Italy.
  Association for Computational Linguistics.

\bibitem[{Liu and Zhou(2020)}]{liu2020introduction}
Zhiyuan Liu and Jie Zhou. 2020.
\newblock Introduction to graph neural networks.
\newblock \emph{Synthesis Lectures on Artificial Intelligence and Machine
  Learning}, 14(2):1--127.

\bibitem[{Loshchilov and Hutter(2018)}]{loshchilov2018decoupled}
Ilya Loshchilov and Frank Hutter. 2018.
\newblock Decoupled weight decay regularization.
\newblock In \emph{International Conference on Learning Representations}.

\bibitem[{Maas et~al.(2013)Maas, Hannun, Ng et~al.}]{maas2013rectifier}
Andrew~L Maas, Awni~Y Hannun, Andrew~Y Ng, et~al. 2013.
\newblock Rectifier nonlinearities improve neural network acoustic models.
\newblock In \emph{Proceedings of the 30th International Conference on Ma-
  chine Learning, Atlanta, Georgia, USA}.

\bibitem[{Manning and Sch{\"u}tze(1999)}]{manning99foundations}
Christopher~D. Manning and Hinrich Sch{\"u}tze. 1999.
\newblock \emph{Foundations of Statistical Natural Language Processing}.
\newblock MIT Press.

\bibitem[{Marcheggiani and Titov(2017)}]{marcheggiani-titov-2017-encoding}
Diego Marcheggiani and Ivan Titov. 2017.
\newblock \href {https://doi.org/10.18653/v1/D17-1159} {Encoding sentences with
  graph convolutional networks for semantic role labeling}.
\newblock In \emph{Proceedings of the 2017 Conference on Empirical Methods in
  Natural Language Processing}, pages 1506--1515, Copenhagen, Denmark.
  Association for Computational Linguistics.

\bibitem[{Mayer and Cysouw(2014)}]{mayer-cysouw-2014-creating}
Thomas Mayer and Michael Cysouw. 2014.
\newblock \href
  {http://www.lrec-conf.org/proceedings/lrec2014/pdf/220_Paper.pdf} {Creating a
  massively parallel {B}ible corpus}.
\newblock In \emph{Proceedings of the Ninth International Conference on
  Language Resources and Evaluation ({LREC}'14)}, pages 3158--3163, Reykjavik,
  Iceland. European Language Resources Association (ELRA).

\bibitem[{McCarthy et~al.(2020)McCarthy, Wicks, Lewis, Mueller, Wu, Adams,
  Nicolai, Post, and Yarowsky}]{mccarthy-etal-2020-johns}
Arya~D. McCarthy, Rachel Wicks, Dylan Lewis, Aaron Mueller, Winston Wu, Oliver
  Adams, Garrett Nicolai, Matt Post, and David Yarowsky. 2020.
\newblock \href {https://aclanthology.org/2020.lrec-1.352} {The {J}ohns
  {H}opkins {U}niversity {B}ible corpus: 1600+ tongues for typological
  exploration}.
\newblock In \emph{Proceedings of the 12th Language Resources and Evaluation
  Conference}, pages 2884--2892, Marseille, France. European Language Resources
  Association.

\bibitem[{Newman(2001)}]{newman2001scientific}
Mark~EJ Newman. 2001.
\newblock Scientific collaboration networks. ii. shortest paths, weighted
  networks, and centrality.
\newblock \emph{Physical review E}, 64(1):016132.

\bibitem[{{\"O}stling and Tiedemann(2016)}]{Ostling2016efmaral}
Robert {\"O}stling and J{\"o}rg Tiedemann. 2016.
\newblock \href {http://ufal.mff.cuni.cz/pbml/106/art-ostling-tiedemann.pdf}
  {Efficient word alignment with {M}arkov {C}hain {M}onte {C}arlo}.
\newblock \emph{Prague Bulletin of Mathematical Linguistics}, 106:125--146.

\bibitem[{Peng et~al.(2018)Peng, Li, He, Liu, Bao, Wang, Song, and
  Yang}]{peng2018large}
Hao Peng, Jianxin Li, Yu~He, Yaopeng Liu, Mengjiao Bao, Lihong Wang, Yangqiu
  Song, and Qiang Yang. 2018.
\newblock Large-scale hierarchical text classification with recursively
  regularized deep graph-cnn.
\newblock In \emph{Proceedings of the 2018 world wide web conference}, pages
  1063--1072.

\bibitem[{Plank and Agi{\'c}(2018)}]{plank-agic-2018-distant}
Barbara Plank and {\v{Z}}eljko Agi{\'c}. 2018.
\newblock \href {https://doi.org/10.18653/v1/D18-1061} {Distant supervision
  from disparate sources for low-resource part-of-speech tagging}.
\newblock In \emph{Proceedings of the 2018 Conference on Empirical Methods in
  Natural Language Processing}, pages 614--620, Brussels, Belgium. Association
  for Computational Linguistics.

\bibitem[{Qi et~al.(2020)Qi, Zhang, Zhang, Bolton, and
  Manning}]{qi-etal-2020-stanza}
Peng Qi, Yuhao Zhang, Yuhui Zhang, Jason Bolton, and Christopher~D. Manning.
  2020.
\newblock \href {https://doi.org/10.18653/v1/2020.acl-demos.14} {{S}tanza: A
  python natural language processing toolkit for many human languages}.
\newblock In \emph{Proceedings of the 58th Annual Meeting of the Association
  for Computational Linguistics: System Demonstrations}, pages 101--108,
  Online. Association for Computational Linguistics.

\bibitem[{Scarselli et~al.(2008)Scarselli, Gori, Tsoi, Hagenbuchner, and
  Monfardini}]{scarselli2008graph}
Franco Scarselli, Marco Gori, Ah~Chung Tsoi, Markus Hagenbuchner, and Gabriele
  Monfardini. 2008.
\newblock The graph neural network model.
\newblock \emph{IEEE transactions on neural networks}, 20(1):61--80.

\bibitem[{Severini et~al.(2022)Severini, Imani, Dufter, and
  Sch{\"u}tze}]{severini2022towards}
Silvia Severini, Ayyoob Imani, Philipp Dufter, and Hinrich Sch{\"u}tze. 2022.
\newblock Towards a broad coverage named entity resource: A data-efficient
  approach for many diverse languages.
\newblock \emph{arXiv preprint arXiv:2201.12219}.

\bibitem[{T{\"a}ckstr{\"o}m et~al.(2013)T{\"a}ckstr{\"o}m, Das, Petrov,
  McDonald, and Nivre}]{tackstrom-etal-2013-token}
Oscar T{\"a}ckstr{\"o}m, Dipanjan Das, Slav Petrov, Ryan McDonald, and Joakim
  Nivre. 2013.
\newblock \href {https://doi.org/10.1162/tacl_a_00205} {Token and type
  constraints for cross-lingual part-of-speech tagging}.
\newblock \emph{Transactions of the Association for Computational Linguistics},
  1:1--12.

\bibitem[{Tiedemann(2012)}]{TIEDEMANN12.463}
Jörg Tiedemann. 2012.
\newblock Parallel data, tools and interfaces in {OPUS}.
\newblock In \emph{Proceedings of the Eight International Conference on
  Language Resources and Evaluation (LREC'12)}, Istanbul, Turkey. European
  Language Resources Association (ELRA).

\bibitem[{Tsai et~al.(2019)Tsai, Riesa, Johnson, Arivazhagan, Li, and
  Archer}]{tsai2019small}
Henry Tsai, Jason Riesa, Melvin Johnson, Naveen Arivazhagan, Xin Li, and Amelia
  Archer. 2019.
\newblock Small and practical bert models for sequence labeling.
\newblock In \emph{Proceedings of the 2019 Conference on Empirical Methods in
  Natural Language Processing and the 9th International Joint Conference on
  Natural Language Processing (EMNLP-IJCNLP)}, pages 3632--3636.

\bibitem[{Turc et~al.(2021)Turc, Lee, Eisenstein, Chang, and
  Toutanova}]{turc2021revisiting}
Iulia Turc, Kenton Lee, Jacob Eisenstein, Ming-Wei Chang, and Kristina
  Toutanova. 2021.
\newblock Revisiting the primacy of {English} in zero-shot cross-lingual
  transfer.
\newblock \emph{arXiv preprint arXiv:2106.16171}.

\bibitem[{Veličković et~al.(2018)Veličković, Cucurull, Casanova, Romero,
  Liò, and Bengio}]{velickovic2018graph}
Petar Veličković, Guillem Cucurull, Arantxa Casanova, Adriana Romero, Pietro
  Liò, and Yoshua Bengio. 2018.
\newblock \href {https://openreview.net/forum?id=rJXMpikCZ} {Graph attention
  networks}.
\newblock In \emph{International Conference on Learning Representations}.

\bibitem[{Wang et~al.(2022)Wang, Ruder, and Neubig}]{wang-etal-2022-expanding}
Xinyi Wang, Sebastian Ruder, and Graham Neubig. 2022.
\newblock \href {https://doi.org/10.18653/v1/2022.acl-long.61} {Expanding
  pretrained models to thousands more languages via lexicon-based adaptation}.
\newblock In \emph{Proceedings of the 60th Annual Meeting of the Association
  for Computational Linguistics (Volume 1: Long Papers)}, pages 863--877,
  Dublin, Ireland. Association for Computational Linguistics.

\bibitem[{Wisniewski et~al.(2014)Wisniewski, P{\'e}cheux, Gahbiche-Braham, and
  Yvon}]{wisniewski2014cross}
Guillaume Wisniewski, Nicolas P{\'e}cheux, Souhir Gahbiche-Braham, and
  Fran{\c{c}}ois Yvon. 2014.
\newblock Cross-lingual part-of-speech tagging through ambiguous learning.
\newblock In \emph{Proceedings of the 2014 Conference on Empirical Methods in
  Natural Language Processing (EMNLP)}, pages 1779--1785.

\bibitem[{Wu et~al.(2020)Wu, Lian, Xu, Wu, and Chen}]{Wu_Lian_Xu_Wu_Chen_2020}
Yongji Wu, Defu Lian, Yiheng Xu, Le~Wu, and Enhong Chen. 2020.
\newblock \href {https://doi.org/10.1609/aaai.v34i01.5455} {Graph convolutional
  networks with markov random field reasoning for social spammer detection}.
\newblock \emph{Proceedings of the AAAI Conference on Artificial Intelligence},
  34(01):1054–1061.

\bibitem[{Yarowsky and Ngai(2001)}]{yarowsky-ngai-2001-inducing}
David Yarowsky and Grace Ngai. 2001.
\newblock \href {https://aclanthology.org/N01-1026} {Inducing multilingual
  {POS} taggers and {NP} bracketers via robust projection across aligned
  corpora}.
\newblock In \emph{Second Meeting of the North {A}merican Chapter of the
  Association for Computational Linguistics}.

\bibitem[{Zeman et~al.(2019)Zeman, Nivre, and Abrams}]{UD}
Daniel Zeman, Joakim Nivre, and Mitchell et~al. Abrams. 2019.
\newblock \href {http://hdl.handle.net/11234/1-3105} {Universal dependencies
  2.5}.
\newblock {LINDAT}/{CLARIAH}-{CZ} digital library at the Institute of Formal
  and Applied Linguistics ({{\'U}FAL}), Faculty of Mathematics and Physics,
  Charles University.

\bibitem[{Zhang et~al.(2018)Zhang, Liu, and Song}]{zhang-etal-2018-sentence}
Yue Zhang, Qi~Liu, and Linfeng Song. 2018.
\newblock \href {https://doi.org/10.18653/v1/P18-1030} {Sentence-state {LSTM}
  for text representation}.
\newblock In \emph{Proceedings of the 56th Annual Meeting of the Association
  for Computational Linguistics (Volume 1: Long Papers)}, pages 317--327,
  Melbourne, Australia. Association for Computational Linguistics.

\end{thebibliography}
\bibliographystyle{acl_natbib}

\appendix

\section{Reproducibility details \label{sec:reprod}}
\begin{table*}[!ht]
	\centering
	\resizebox{1\textwidth}{!}{%
		\begin{tabular}{lrrrrrrrrrrrrrrrrrrrrr}
			\toprule
			
			& afr & amh & eus & bul & hin & ind & lit & pes & por & tel & tur & bel & mar & AVG && bam & myv & glv & yor & AVG \\  \cmidrule{2-15} \cmidrule{17-21} 
			with XLM-R  & 87.7 & 82.4   & 70.9    &  90.1   &  81.8   & 85.3 &  85.7   &  81.8  &  89.2  &  83.8   &  80.1 & 85.9    &   87.9 & \textbf{84.1} &
			&  43.0  & 55.2  & 50.0 & 39.0  & 46.8 \\ 
			
			without XLM-R & 88.4 & 82.8 & 72.7 &89.3 & 73.7 & 80.2 & 83.9 & 71.3 & 85.0 & 80.1 & 77.8 & 85.2 & 82.0 & 81.0 & &
			65.4    & 64.4 & 63.9    &  59.9 & \textbf{63.4} \\
			\bottomrule
		\end{tabular}
	}
	\caption{Accuracy on UD v2.10
		for \ourmethodtwo 	 when transferring from all
		training languages	(i.e., \ourmethodtwo-All) with and without using XLM-R for the transfer in \ourmethodtwo. 
		\label{tab:app_main}}
\end{table*}

\subsection{Data editions}\label{appendix_data}

Table~\ref{tab:versions} lists the PBC editions used for all the experiments in this paper.

\begin{table}[!h]
	\centering
	\small
	\resizebox{.5\textwidth}{!}{%
		\begin{tabular}{lll|ll}
			\toprule
			&Lang    & Edition & Lang  &  Edition   \\ \midrule
			& Arabic  & arb-x-bible        &  Hungarian	 &  hun-x-bible-newworld \\
			& Chinese & zho-x-bible-newworld & Afrikansc & afr-x-bible-newworld  \\
			& Danish & dan-x-bible-newworld & Amharic	& amh-x-bible-newworld \\
			& English* & eng-x-bible-mixed & Basque & eus-x-bible-navarrolabourdin \\
			& Finnish* &  fin-x-bible-helfi   & Belarusian &  bel-x-bible-bokun  \\
			& French & fra-x-bible-louissegond &  Bulgarian & bul-x-bible-newworld \\
			& German & deu-x-bible-bolsinger & Hindi  & hin-x-bible-bsi \\
			& Irish 	& gle-x-bible & Indonesian & ind-x-bible-newworld \\
			& Italian 	& ita-x-bible-2009 &  Lithuanian & lit-x-bible-ecumenical \\
			& Polish 	& pol-x-bible-newworld & Marathi & mar-x-bible  \\
			& Russian & rus-x-bible-newworld    & Persian &  pes-x-bible-newmillennium2011      \\
			& Spanish & spa-x-bible-newworld   & Portuguese  &  por-x-bible-newworld1996      \\
			& Swedish & swe-x-bible-newworld     & Telugu &  tel-x-bible     \\
			& Tamil   & tam-x-bible-newworld         & 	Turkish  &  tur-x-bible-newworld   \\ 
			& Urdu   & urd-x-bible-2007 &  Bambara   &  bam-x-bible    \\ 
			& Czech & ces-x-bible-newworld & Erzya & myv-x-bible \\
			& Greek   & ell-x-bible-newworld     & Manx &  glv-x-bible      \\
			& Hebrew* & heb-x-bible-helfi & Yoruba & yor-x-bible-2010 \\
 \bottomrule
			
		\end{tabular}
	}
	\caption{PBC editions for all used languages. *Edition from \citet{imani-etal-2022-graph}.
		\label{tab:versions}}
\end{table}

\subsection{Universal Dependency tests specification}\label{appendix_ud}

Table~\ref{tab:ud_test_version} lists the Universal Dependency testsets used in our experiments.

\begin{table}[!h]
	\centering
	\small
		\begin{tabularx}{\columnwidth}{XX}
			\toprule
			Lang    & Test   \\ \midrule
			 Afrikaans & af\_afribooms-ud-test  \\
			 Amharic	& am\_att-ud-test \\
			 Basque & eu\_bdt-ud-test \\
			 Belarusian &  be\_hse-ud-test \\
			 Bulgarian & bg\_btb-ud-test \\
			 Hindi  & hi\_hdtb-ud-test \\
			 Ind & id\_gsd-ud-test \\
			 Lithuanian & lt\_alksnis-ud-test \\
			 Marathi & mr\_ufal-ud-test.  \\
			 Persian &  fa\_seraji-ud-test    \\
			 Portuguese  &  pt\_bosque-ud-test    \\
			Telugu &  te\_mtg-ud-test    \\
			Turkish  &  tr\_imst-ud-test   \\ 
			 Bambara   &  bm\_crb-ud-test    \\ 
			 Erzya & myv\_jr-ud-test \\
			 Manx &  gv\_cadhan-ud-test    \\
			 Yoruba & yo\_ytb-ud-test \\
			\bottomrule
			
		\end{tabularx}
%	}
	\caption{Universal Dependency test sets used in our experiments.
		\label{tab:ud_test_version}}
\end{table}

\subsection{Models parameters}\label{app:model_param}

\paragraph{\ourmethod} The \ourmethod{} is implemented using the PyTorch geometric library.\footnote{\url{https://pytorch-geometric.readthedocs.io/en/latest/}}
All hyperparameters are tuned on the dev set. 
\ourmethodone{} has 2 layers of MLP of size 2048 while \ourmethodtwo{}
uses four layers of transformer with hidden size 2048 and 16 attention heads.
Although we didn't observe a difference between different sizes from 512 to 2048.
We tuned the learning rate, batch size, and $\gamma$ (the self-learning threshold)
over the validation languages. In \ourmethodone{} learning rate and 
batch size are respectively 
$0.001$, $8$, and in \ourmethodtwo{} $0.00001$, and $32$. In general, when using
XLM-R embeddings, the model has higher confidence, so the $\gamma$ parameter is set 
to $0.95$ when not using XLM-R embeddings and $0.98$ when using XLM-R embeddings. 
The whole model needs about $16 GB$ of GPU memory. \ourmethodone{}
takes about 2 hours to train and \ourmethodtwo about 12 hours. We used early stopping
with patience of 8 for both \ourmethodone{} and \ourmethodtwo{}.

\paragraph{Neural POS tagger}
We run our method on up to 48 cores of Intel(R)
Xeon(R) CPU E7-8857 v2 with 1TB memory and a
single GeForce GTX 1080 GPU with 8GB memory.
The POS tagger uses the Flair framework \citep{akbik-etal-2019-flair}
and SequenceTagger model with 128 hidden size, the  "xlm-roberta-base" embeddings, and AdamW optimizer \citet{loshchilov2018decoupled}.
The hyperparameters, including the fixed number of epochs (15)
are tuned using the UD development sets of the development languages.
Each Neural POS tagger was trained in less than 30 minutes.

\section{Contextualized vs. Static embeddings }\label{appendix_results}

Table \ref{tab:app_main} shows results obtained with our \ourmethodtwo with and without using XLM-R embeddings for projection. Note that the final neural POS tagger models always use XLM-R embeddings, even for languages unseen during XLM-R pretraining.

\end{document}